\documentclass[runningheads]{llncs}
\usepackage[T1]{fontenc}

\usepackage{amsmath,amssymb,amsfonts}
\usepackage{graphicx}
\usepackage{subcaption}
\usepackage{float}
\usepackage{amsmath}
\usepackage{booktabs}

\usepackage{graphicx}
\usepackage{todonotes}
\usepackage{soul}
\usepackage{tikz}
\usetikzlibrary{arrows.meta}
\usetikzlibrary{calc}
\usepackage{hyperref}
\usetikzlibrary{positioning,arrows.meta,calc,fit}
\usepackage[export]{adjustbox} 

\usepackage{ragged2e}
\usepackage[utf8]{inputenc}
\usepackage[T1]{fontenc}
\DeclareUnicodeCharacter{0394}{$\Delta$}
\DeclareUnicodeCharacter{2212}{-} 

\usepackage{graphicx,verbatim}
\usepackage{multirow}
\begin{document}
%

\title{MSA-DCNN: A Data-Efficient Multi-Scale Attention Deformable CNN for Medical Image Classification}
\titlerunning{MSA-DCNN}
\author{Hamza Hussaini \inst{1}\orcidID{0009-0005-6463-1607} \and
Shahana Bano\inst{1}\orcidID{0000-0002-6907-1777} \and
Eyad Elyan\inst{1}\orcidID{0000-0002-8342-9026}
\and
Carlos Francisco Moreno-García\inst{1}\orcidID{0000-0001-7218-9023}}
\authorrunning{Hussaini et al.}
%
\institute{Robert Gordon University, School of Computing, Engineering and Technology, Aberdeen, Scotland, UK\\
\email{h.hussaini;s.bano;e.elyan;c.moreno-garcia@rgu.ac.uk}\\
\url{http://www.rgu.ac.uk}}

\maketitle              
\begin{abstract}
Existing deep learning methods perform well in medical image classification but struggle with multi-scale morphology and limited annotations due to fixed sampling and data-hungry training. Existing approaches address these challenges in isolation: DCN-based models provide adaptive sampling but lack explicit multi-scale attention fusion and label-efficient regularisation; multi-scale architectures typically rely on static fusion; and semi-supervised methods target label scarcity without jointly modelling adaptive cross-scale representations. We propose MSA-DCNN, a scale-consistent deformable attention learning framework that introduces adaptive multi-scale sampling, within-scale saliency refinement, learned cross-scale fusion, and auxiliary self-distillation within a unified optimisation scheme, with potential to generalise to structurally heterogeneous anatomy. We evaluate on three public benchmarks and an external hold-out set for leukaemia. MSA-DCNN demonstrates competitive and often better performance against ViT baselines, CNN baselines, a deformable CNN baseline, and a MICCAI semi-supervised baseline under distribution shift and label scarcity in accuracy, F1, and AUC (binary), while using fewer parameters. Ablations confirm complementary component contributions, supporting MSA-DCNN as a practical foundation for data-efficient medical image classification.


\end{abstract}
\section{Introduction}

Automated medical image classification is a key enabler of computational healthcare, delivering fast, objective, and reproducible analysis across radiography, histopathology, and microscopy \cite{khalifa2024ai,obuchowicz2024clinical}. Convolutional neural networks (CNNs) remain the workhorse due to their hierarchical feature learning and strong performance across lung, brain, oncology, and haematology tasks \cite{yu2025small,guo2025artificial,rayed2024deep,mall2023comprehensive,mienye2025deep}. Yet, standard CNNs inherit fixed receptive fields and uniform sampling, which are often poorly matched to the heterogeneous textures and multi-scale structural variability of medical imagery \cite{yu2021convolutional,tong2024hybrid,wei2021fine}. Prior remedies, including attention modules and multi-scale designs, improve feature focus and coverage \cite{subba2025computationally,han2024dm,jin2023simplified,zhang2022rcmnet}, but often rely on global aggregation that can wash out subtle cues \cite{hussaini2025modified}, retain fixed kernels that do not adapt sampling across scales \cite{xu2022isanet}, and rarely provide internal supervision to strengthen shallow layers in data-limited regimes \cite{kishore2025designing}. Moreover, medical image classification is frequently constrained by limited labelled data \cite{ju2024universal,kunanbayev2024training,li2024semi,chen2022recent}, limiting generalisation and sensitivity to minority patterns.

We address these gaps with the Multi-Scale Attention Deformable Convolutional Neural Network (MSA-DCNN). This work introduces a general principle for scale-consistent learning under structural heterogeneity and label scarcity, designed to perform effectively at low label fractions with a lean parameter budget while enforcing semantic alignment across resolution levels. Critically, attention is applied scale-specifically, and the resulting features are fused by a learned multi-scale attention mechanism, coupling within-scale recalibration with across-scale selection, an integration not explicitly studied in prior deformable or attention-based CNNs.

We evaluate MSA-DCNN on three publicly available benchmarks against data-efficient baselines spanning diverse morphological patterns, from dermoscopic imagery \cite{adepu2023melanoma} to fine-grained cellular structure in blood and leukaemia smears \cite{acevedo2020dataset,gupta2022c}. Ablations confirm complementary component effects, and external testing on an independent leukaemia cohort assesses generalisation. Across benchmarks, MSA-DCNN improves AUC (binary), accuracy, and F1 with fewer parameters.

\textbf{Contributions:}
We introduce a scale-consistent learning framework for multi-scale medical image analysis, implemented as a compact CNN with three established principles:
\begin{enumerate}
    \item Adaptive receptive-field optimisation, where deformable sampling is structured at each scale to preserve fine-grained detail while maintaining global context.
    \item Scale-consistent saliency refinement, achieved through sub-block-pooled channel and spatial attention, which regularises within-scale feature distributions under limited supervision.
    \item Content-aware cross-scale integration via learned attention over projected multi-resolution embeddings, with auxiliary self-distillation to align shallow and deep representations for stable, label-efficient learning.
\end{enumerate}

Unlike prior multi-scale or deformable CNNs that treat adaptive sampling, attention, and fusion as separate stages, MSA-DCNN jointly learns where to sample, what to emphasise, and how to combine features under a unified objective. Attention regularises scale-local representations within each deformable branch, while a learned multi-scale operator performs content-adaptive selection across projected embeddings (Eq.~\ref{eq:Us}--\ref{eq:msa}). An auxiliary self-distillation head enforces cross-scale semantic alignment via KL-divergence minimisation (Eq.~\ref{eq:sd}), promoting coherent representation geometry and explaining the monotonic gains observed in the ablation study.

\section{Methodology}\label{methods}  

\begin{figure}[t]
\centering
\captionsetup[subfigure]{labelformat=parens,labelsep=space,justification=centering,font=small}
\begin{adjustbox}{max width=\linewidth, max totalheight=0.5\textheight}
\begin{minipage}{0.98\linewidth}

  \begin{subfigure}[t]{\linewidth}
    \centering
    \includegraphics[width=\linewidth]{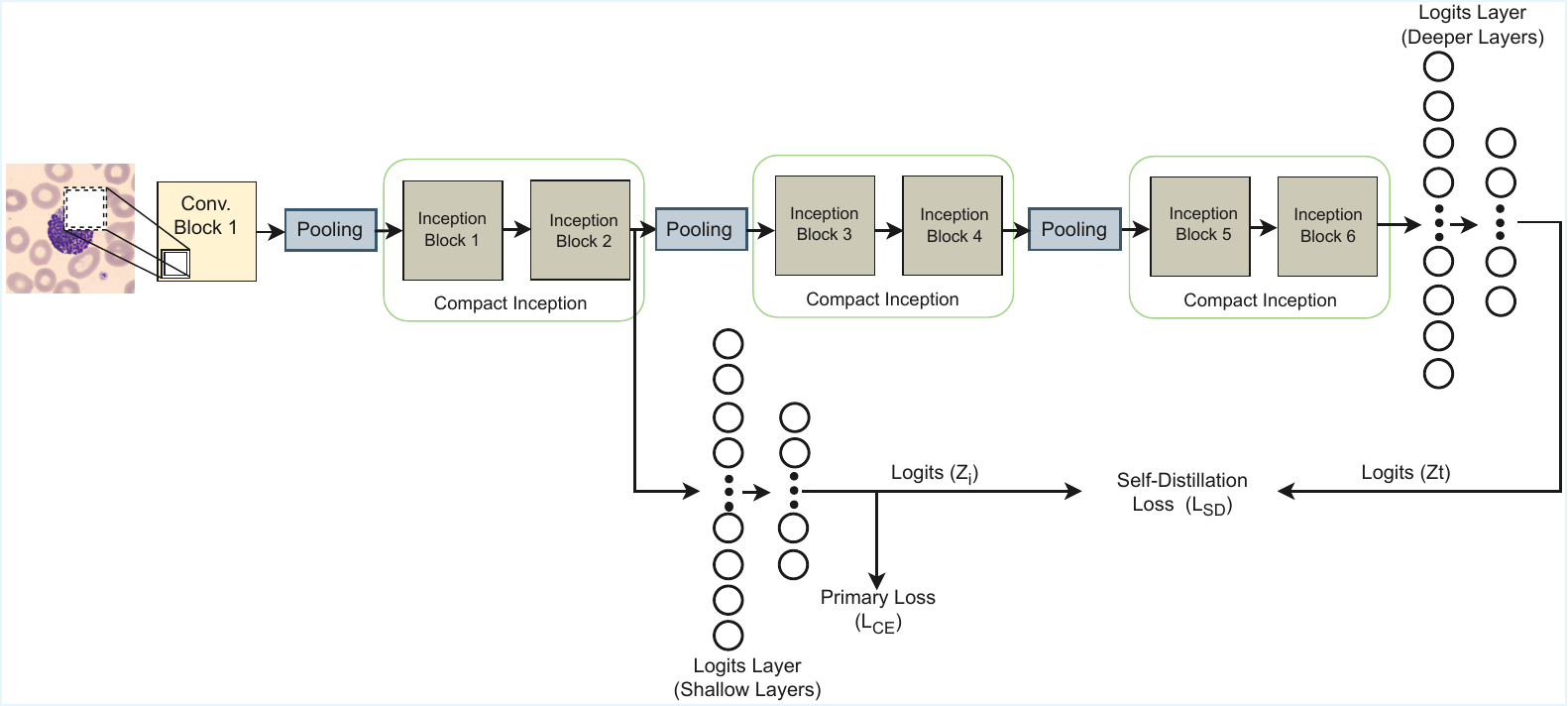}
    \caption{Pipeline overview.}
    \label{fig:overview}
  \end{subfigure}

  \vspace{0.35em}

  \begin{subfigure}[t]{0.49\linewidth}
    \centering
    \includegraphics[width=\linewidth]{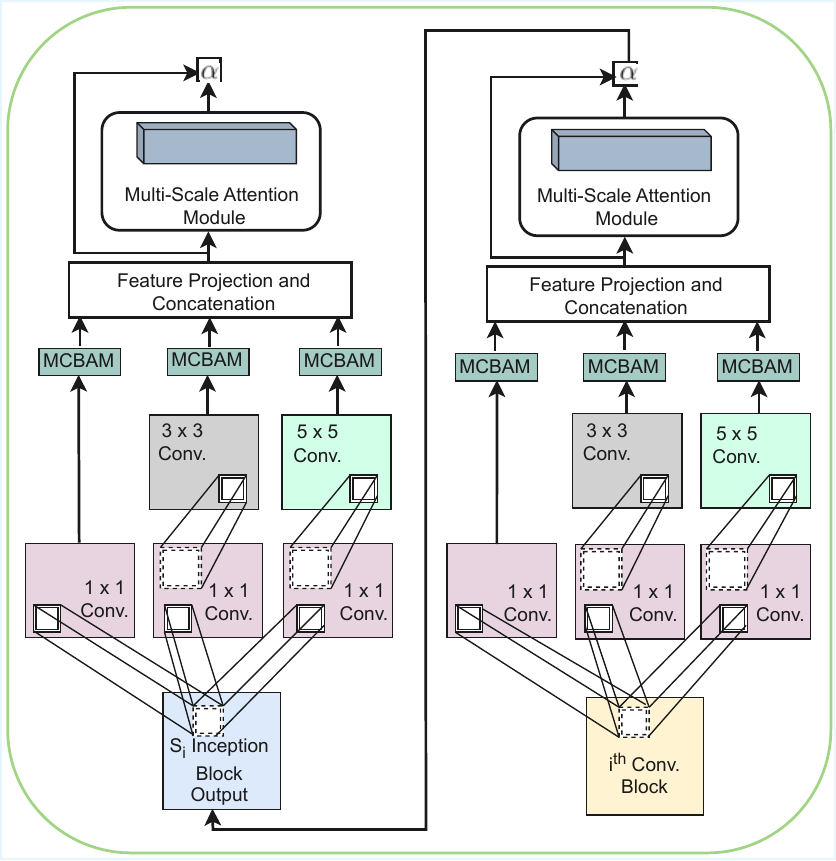}
    \caption{Compact Inception.}
    \label{fig:zoom-msdconv}
  \end{subfigure}\hfill
  \begin{subfigure}[t]{0.49\linewidth}
    \centering
    \includegraphics[width=\linewidth]{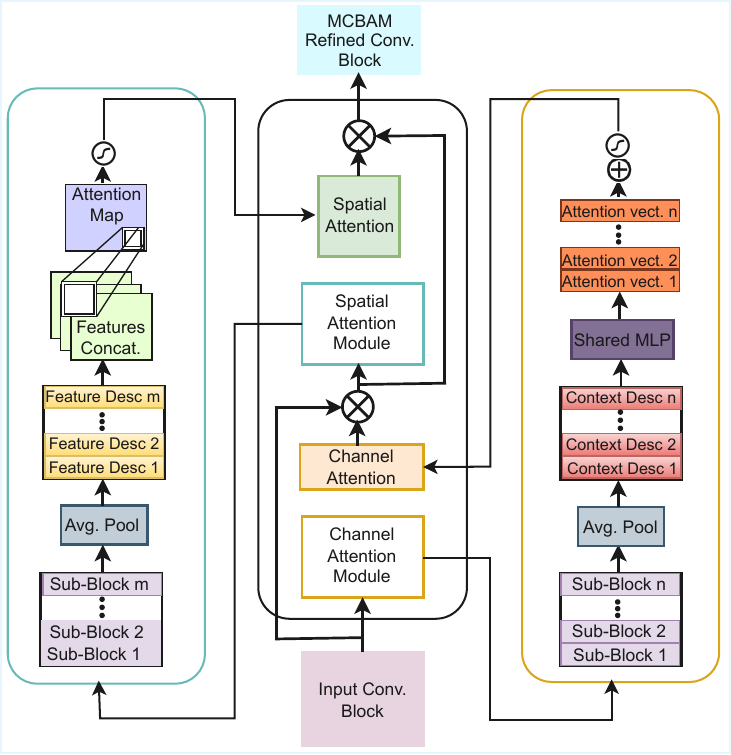}
    \caption{MCBAM.}
    \label{fig:zoom-mcbam}
  \end{subfigure}

\end{minipage}
\end{adjustbox}
\caption{MSA-DCNN overview and key components: (a) end-to-end pipeline; (b) compact Inception (multi-scale deformable conv); (c) MCBAM internals.}
\label{fig:msadcnn_full}
\end{figure}

Figure~\ref{fig:overview} provides a concise overview of the proposed MSA-DCNN, which is designed around jointly optimising \emph{where} features are sampled, \emph{how} saliency is preserved within each scale, and \emph{how} information is integrated across scales under limited supervision. After standard preprocessing and a lightweight stem, features traverse three compact Inception stages detailed in Figure~\ref{fig:zoom-msdconv}. Each stage comprises two sequential Inception blocks with three parallel branches operating at distinct receptive fields, where deformable convolutions adapt the sampling grid to structurally heterogeneous, variable-scale patterns. Within each branch, a scale-specific MCBAM with sub-block pooling (shown in Figure~\ref{fig:zoom-mcbam}) performs channel--spatial recalibration that preserves fine-grained morphological detail. The branch outputs are then projected to a common resolution (via $1\times1$ conv + resize), and a multi-scale attention module computes content-dependent weights across scales to yield a unified representation. An auxiliary head attached to the first stage supplies self-distillation during training, aligning shallow and deep embeddings, while the main path proceeds to global average pooling, dropout, and a linear classification head. The model is trained end-to-end and uses a single forward pass at inference.

\subsection{Multi-Scale Deformable Convolution with Attention Fusion}

\noindent Let $s \in \{1,\dots,S\}$ index the deformable convolution branches (in this case, $S=3$). The output $F$ of branch $s$ is defined as:
\begin{equation}
F_s \in \mathbb{R}^{C_s \times H_s \times W_s}.
\label{eq:Fs}
\end{equation}
\noindent where $C$, $H$ and $W$ represent the channel, height, and width of the output.

\subsubsection{Deformable Convolution}

\noindent Deformable convolution introduces learnable offsets $\Delta p_k$ for each sampling point \cite{dai2017deformable}. The output at position $p_0$ is: 
\begin{equation}
\mathrm{DefConv}(p_0)
=
\sum_{k=1}^{K}
w(k)\,
x\!\left(p_0 + p_k + \Delta p_k\right),
\label{eq:defconv}
\end{equation}
where $p_k$ denotes the fixed grid offsets and $w(k)$ are the kernel weights. The offsets $\Delta p_k$ allow the receptive field to shift adaptively, enabling improved alignment with structurally heterogeneous and multi-scale patterns.

\subsubsection{Scale-Specific MCBAM Refinement}

Each branch is refined using a two-stage MCBAM module. First, channel attention $A^{(c)}_s$ is computed from $F_s$ as:

\begin{equation}
A^{(c)}_s(F_s)
=
\sigma\!\left(
\sum_{n=1}^{N}
W_{ho,s}\,
\delta\!\left(
W_{ih,s}\,
\mathrm{AvgPool}_{h_s \times w_s \times 1}(P_{s,n})
\right)
\right)
\in \mathbb{R}^{1 \times 1 \times C_s},
\label{eq:channelAtt}
\end{equation}
where $W_{ih,s}$ and $W_{ho,s}$ are MLP learnable weights, $\delta$ is the ReLU activation, $N$ is the number of spatial sub-blocks (4 in this case, which improves computational efficiency), and $\sigma$ is the sigmoid activation. Channel refinement is then applied as:
\begin{equation}
F_s^{(c)} = A^{(c)}_s(F_s) \odot F_s.
\label{eq:channelRefined}
\end{equation}

The channel-refined map $F_s^{(c)}$ is partitioned into ($m$ = 3, which improves efficiency) channel sub-blocks 
$\{Q_{s,j}\}_{j=1}^{m}$, as shown in the spatial attention component of Figure~\ref{fig:zoom-mcbam}.  
Spatial attention is computed by aggregating sub-block descriptors and applying a convolutional filter ($7\times7$ spatial dimension is used, as derived from \cite{hussaini2025modified}):

\begin{equation}
\begin{aligned}
A^{(s)}_s(F_s^{(c)}) &=
\sigma\!\left(
\mathrm{Conv}_{7\times7}\!
\left(
\mathrm{Conc.}\!\left[
\mathrm{AP}_{1\times1\times d}(Q_{s,1}),
\dots,
\mathrm{AP}_{1\times1\times d}(Q_{s,m})
\right]
\right)
\right) \\
&\in \mathbb{R}^{1 \times H_s \times W_s}.
\end{aligned}
\label{eq:spatialAtt}
\end{equation}

Finally, spatial attention is applied to the channel-refined feature map to obtain the MCBAM-modified output:
\begin{equation}
F_s' = A^{(s)}_s(F_s^{(c)}) \odot F_s^{(c)}.
\label{eq:mcbamFinal}
\end{equation}

\subsubsection{Projection and Multi-Scale Attention Fusion}

\noindent Each refined feature map is projected into a common space using a scale-projection operator:
\begin{equation}
U_s(F_s') = 
\mathrm{resize}\!\left(
\mathrm{Conv}_{1\times1}(F_s')
\right)
\in \mathbb{R}^{C \times H \times W}.
\label{eq:Us}
\end{equation}

Sample-wise multi-scale attention is computed from concatenated global descriptors as:
\begin{equation}
\alpha =
\mathrm{softmax}
\!\left(
W_2\,
\delta\!\left(
W_1\,
\mathrm{Concat}_{s=1}^{S}
\left[
\mathrm{GAP}(U_s(F_s'))
\right]
\right)
\right)
\in \mathbb{R}^{S}.
\label{eq:alpha}
\end{equation}

\noindent $W_1$ and $W_2$ are learnable parameters for multi-scale attention; GAP denotes global average pooling. The fused multi-scale representation is:

\begin{equation}
F_{\mathrm{MSA}}
=
\sum_{s=1}^{S}
\alpha\,U_s(F_s').
\label{eq:msa}
\end{equation}

Eq.~(\ref{eq:Fs}--\ref{eq:msa}) define a scale-attentive operator that jointly regularises within-scale saliency and across-scale selection, yielding a unified, content-adaptive embedding for irregular morphology.

\subsection{Self-Distillation}

We formalise self-distillation as a geometric constraint on cross-scale representation alignment that regularises early deformable offsets by aligning low- and high-resolution semantic manifolds in low-label regimes. This enforces invariance of semantic structure across resolution levels rather than simple logit matching. Let $z_T$ and $z_i$ denote deep and shallow projections, respectively. The distillation loss is:

\begin{equation}
L_{\mathrm{SD}}
=
\mathrm{KLDiv}
\!\left(
\mathrm{Softmax}\!\left(\frac{z_T}{\tau}\right),
\,
\mathrm{Softmax}\!\left(\frac{z_i}{\tau}\right)
\right),
\label{eq:sd}
\end{equation}
where $\tau$ is a temperature parameter.  
The overall training objective is
\begin{equation}
L_{\mathrm{total}}
=
\lambda_{CE}L_{\mathrm{CE}}
+
\lambda_{FL}L_{\mathrm{Focal}}
+
\lambda L_{\mathrm{SD}},
\label{eq:total_loss}
\end{equation}

where $L_{\mathrm{CE}}$ denotes the class-weighted cross-entropy loss, $L_{\mathrm{Focal}}$ denotes the focal loss for handling class imbalance, and $L_{\mathrm{SD}}$ represents the self-distillation loss, and the $\lambda_{CE}$, $\lambda_{FL}$, and $\lambda$ are the corresponding loss weighting coefficients.

\section{Experiments}

\subsection{Datasets and Evaluation Metrics}

We evaluate MSA\text{-}DCNN on three public benchmarks and an external hold-out set. Our experiments assess quality, robustness, and label efficiency of multi-scale representations under distribution shift and class imbalance. \textbf{C\text{-}NMC} contains 15{,}114 white blood cell smear images ($450{\times}450$, RGB) with class imbalance between normal and leukaemia cells; \textbf{PBC} comprises 17{,}092 images across eight peripheral blood cell classes ($360{\times}363$, RGB) with pronounced class imbalance; and \textbf{ISIC\text{-}2020} is a large-scale dermoscopic dataset with severe skew ($\approx$32k benign vs.\ 500 malignant) and varying image resolutions. For external validation, we evaluate on an entirely independent leukaemia smear dataset\footnote{https://www.kaggle.com/datasets/mehradaria/leukemia?resource=download}, where pre-, early-, and pro-leukaemia labels are merged into a single leukaemia class to match the binary setting. Importantly, this external dataset is \emph{not used at any stage of model development}, including training, validation, hyperparameter tuning, model selection, or internal testing. Instead, it is reserved exclusively for post hoc evaluation to assess the generalisation capability and robustness of MSA-DCNN under distribution shift, a common challenge in real-world clinical deployment. We report \textbf{Accuracy} and \textbf{F1-score} for all datasets, and \textbf{AUC} for binary tasks. Model efficiency is assessed using \textbf{parameter count} and floating-point operations \textbf{(FLOPs)}.

\subsection{Experimental Details}

Experiments were conducted on an NVIDIA A100 GPU using PyTorch with a 70/15/15 train/validation/test split. Images were processed at dataset-specific resolutions: PBC ($299{\times}299$), C\text{-}NMC ($450{\times}450$), and ISIC-2020 ($768{\times}768$). All methods, including a DeiT ImageNet-pretrained transformer, followed identical protocols and were trained with Adam ($1{\times}10^{-4}$, batch size 32) for 20 epochs using class-balanced sampling, class-weighted cross-entropy (binary BCE for two-class tasks), focal loss, and minority-targeted augmentations. Evaluation used normalised test images only; binary tasks employed BCE-with-logits, and PBC used categorical cross-entropy. For MSA-DCNN, the auxiliary self-distillation head used a temperature parameter of $\tau=4$ and a loss weight of $\lambda=0.3$, which were fixed across all datasets. Hyperparameters were selected on the validation set and kept unchanged across datasets. Results are reported as mean $\pm$ standard deviation over five random seeds, with significance assessed using paired Wilcoxon tests ($p<0.05$) and 1,000-sample bootstrap confidence intervals. Code will be released upon acceptance.

\subsection{Quantitative Performance Comparison}

Table~\ref{table:CNN} shows that MSA-DCNN achieves competitive and often better performance across datasets, surpassing foundational and data-efficient transformers (DINOv2-S \cite{oquab2023dinov2}, DeiT-S \cite{touvron2021training}), compact CNNs (Inception-v3 \cite{szegedy2016rethinking}, EfficientNet \cite{tan2019efficientnet}), the deformable-convolution-based DGConv \cite{gong2021deformable}, and a semi-supervised baseline (TS-MS \cite{li2024semi}) while using substantially fewer parameters. DGConv consistently provides the strongest baseline performance across most datasets, highlighting the effectiveness of deformable convolutions for modelling geometric variations; however, MSA-DCNN achieves further gains in AUC, accuracy, and F1-score with lower parameter complexity. As shown in Fig.~\ref{fig:subset-pbc}, ~\ref{fig:subset-cnmc}, and ~\ref{fig:subset-isic}, performance remains higher under reduced label fractions, indicating improved label efficiency relative to both supervised and semi-supervised baselines. Gains in AUC and F1 persist under class imbalance, reflecting improved sensitivity to minority patterns. Paired Wilcoxon tests across seeds confirm significant AUC gains (C-NMC: $p=0.018$, ISIC-2020: $p=0.011$, ALL: $p=0.024$), with 95\% bootstrap CIs of [+0.014, +0.029], [+0.010, +0.030], and [+0.006, +0.032], respectively. Despite increased FLOPs at higher resolutions, MSA-DCNN remains computationally competitive, and performance on the external ALL cohort degrades modestly yet remains consistently above all baselines, supporting robustness under distribution shift.

\begin{table}[htbp!]
  \caption{Performance and efficiency of MSA-DCNN and baselines on PBC, C-NMC, ISIC-2020, and an external ALL set. For MSA-DCNN, values denote mean $\pm$ std; $^{\dagger}$ indicates significant AUC gains ($p<0.05$)}
  \label{table:CNN}
  \centering
  \begingroup
  \setlength{\tabcolsep}{2pt}
  \renewcommand{\arraystretch}{0.95}
  \fontsize{8}{9}\selectfont
  \begin{tabular}{llccccc}
    \hline
    \textbf{DB} & \textbf{Method} & \textbf{AUC} & \textbf{Acc} & \textbf{F1} & \textbf{FLOPs} & \textbf{Params} \\
    \hline
    \multirow{6}{*}{PBC}
        & Inception-v3 \cite{szegedy2016rethinking} & -- & 0.92 & 0.91 & 5.74 & 23.9 \\
        & EfficientNet-B3 \cite{tan2019efficientnet} & -- & 0.90 & 0.89 & \textbf{1.78} & 12.0 \\
        & DeiT-S \cite{touvron2021training}  & --   & 0.89 & 0.88 & 5.03 & 22.1 \\
        & DINOv2-S \cite{oquab2023dinov2}        & --   & 0.91 & 0.90 & 11.08 & 21.3 \\
        & TS-MS \cite{li2024semi}          & -- & 0.92 & 0.91 & 3.10 & 8.4 \\
        & DGConv \cite{gong2021deformable}          & -- & 0.92 & 0.92 & 3.42 & 7.98 \\
        & \textbf{MSA-DCNN}               & -- & \textbf{0.94} & \textbf{0.93} & 1.99 & \textbf{5.25} \\
    \hline
    \multirow{6}{*}{C-NMC}
        & Inception-v3 \cite{szegedy2016rethinking} & 0.94 & 0.93 & 0.92 & 7.57 & 23.9 \\
        & EfficientNet-B3 \cite{tan2019efficientnet} & 0.92 & 0.91 & 0.90 & \textbf{3.12} & 12.0 \\
        & DeiT-S \cite{touvron2021training}  & 0.91 & 0.90 & 0.89 & 6.99 & 22.1 \\
        & DINOv2-S \cite{oquab2023dinov2}        & 0.93 & 0.92 & 0.90 & 13.97 & 21.3 \\
        & TS-MS \cite{li2024semi}          & 0.91 & 0.90 & 0.88 & 4.60 & 8.4 \\
        & DGConv \cite{gong2021deformable}          & 0.94 & 0.92 & \textbf{0.93} & 4.54 & 7.98 \\
        & \textbf{MSA-DCNN}               & \textbf{0.96$\pm$0.003}$^{\dagger}$ & \textbf{0.94$\pm$0.002} & \textbf{0.93$\pm$0.004} & 3.80 & \textbf{5.25} \\
    \hline
    \multirow{6}{*}{ISIC-2020}
        & Inception-v3 \cite{szegedy2016rethinking} & 0.95 & 0.94 & 0.93 & 16.60 & 23.9 \\
        & EfficientNet-B3 \cite{tan2019efficientnet} & 0.93 & 0.92 & 0.91 & \textbf{10.98} & 12.0 \\
        & DeiT-S \cite{touvron2021training}  & 0.92 & 0.91 & 0.89 & 13.40 & 22.1 \\
        & DINOv2-S \cite{oquab2023dinov2}        & 0.94 & 0.93 & 0.91 & 19.04 & 21.3 \\
        & TS-MS \cite{li2024semi}          & 0.95 & 0.93 & 0.92 & 12.10 & 8.4 \\
        & DGConv \cite{gong2021deformable}          & 0.96 & 0.94 & \textbf{0.94} & 9.36 & 7.98 \\
        & \textbf{MSA-DCNN}               & \textbf{0.97$\pm$0.003}$^{\dagger}$ & \textbf{0.95$\pm$0.002} & \textbf{0.94$\pm$0.006} & 11.07 & \textbf{5.25} \\
    \hline
    \multirow{6}{*}{\begin{tabular}{@{}c@{}}Hold-Out\\ALL\end{tabular}}
        & Inception-v3 \cite{szegedy2016rethinking} & 0.91 & 0.90 & 0.89 & 7.57 & 23.9 \\
        & EfficientNet-B3 \cite{tan2019efficientnet} & 0.89 & 0.89 & 0.88 & \textbf{3.12} & 12.0 \\
        & DeiT-S \cite{touvron2021training}  & 0.88 & 0.88 & 0.87 & 6.24 & 22.1 \\
        & DINOv2-S \cite{oquab2023dinov2}        & 0.90 & 0.90 & 0.88 & 13.97 & 21.3 \\
        & TS-MS \cite{li2024semi}          & 0.89 & 0.87 & 0.86 & 4.60 & 8.4 \\
        & DGConv \cite{gong2021deformable}          & 0.91 & 0.90 & 0.91 & 4.54 & 7.98 \\
        & \textbf{MSA-DCNN}               & \textbf{0.93$\pm$0.005}$^{\dagger}$ & \textbf{0.92$\pm$0.004} & \textbf{0.91$\pm$0.006} & 3.80 & \textbf{5.25} \\
    \hline
  \end{tabular}
  \endgroup
\end{table}

\subsection{Ablation Studies}

\begin{table}[htbp!]
  \caption{Full Ablation Studies on the PBC Dataset, and Reduced Ablation Studies on the C-NMC and ISIC-2020 Datasets.}
  \label{table:ablation}
  \centering
  \small
  \begin{tabular}{lcccccccc}
    \toprule
    Dataset & Baseline & DC & SSA & MSA & SD & AUC & Acc & $F_1$ \\
    \midrule
    \textbf{PBC}
        & \checkmark &            &            &            &            & -- & 0.88 & 0.88 \\
        & \checkmark & \checkmark &            &            &            & -- & 0.90 & 0.90 \\
        & \checkmark &            & \checkmark &            &            & -- & 0.90 & 0.89 \\
        & \checkmark &            &            & \checkmark &            & -- & 0.91 & 0.91 \\
        & \checkmark &            &            &            & \checkmark & -- & 0.91 & 0.90 \\

        & \checkmark & \checkmark & \checkmark &            &            & -- & 0.92 & 0.92 \\
        & \checkmark & \checkmark &            & \checkmark &            & -- & 0.92 & 0.91 \\
        & \checkmark & \checkmark &            &            & \checkmark & -- & 0.92 & 0.91 \\

        & \checkmark &            & \checkmark & \checkmark &            & -- & 0.92 & 0.92 \\
        & \checkmark &            & \checkmark &            & \checkmark & -- & 0.92 & 0.91 \\
        & \checkmark &            &            & \checkmark & \checkmark & -- & 0.92 & 0.91 \\

        & \checkmark & \checkmark & \checkmark & \checkmark &            & -- & 0.93 & \textbf{0.93} \\
        & \checkmark & \checkmark & \checkmark &            & \checkmark & -- & 0.93 & \textbf{0.93} \\
        & \checkmark & \checkmark &            & \checkmark & \checkmark & -- & 0.93 & 0.92 \\
        & \checkmark &            & \checkmark & \checkmark & \checkmark & -- & 0.93 & \textbf{0.93} \\

        & \checkmark & \checkmark & \checkmark & \checkmark & \checkmark & -- &
          \textbf{0.94} & \textbf{0.93} \\
    \midrule
    C\text{-}NMC
        & \checkmark &            &            &            &            & 0.92 & 0.91 & 0.90 \\
        & \checkmark & \checkmark &            &            &            & 0.94 & 0.92 & 0.92 \\
        & \checkmark & \checkmark & \checkmark &            &            & 0.95 & 0.93 & 0.92 \\
        & \checkmark & \checkmark & \checkmark & \checkmark &            & 0.95 & 0.93 & 0.92 \\
        & \checkmark & \checkmark & \checkmark & \checkmark & \checkmark & \textbf{0.96} &
          \textbf{0.94} & \textbf{0.93} \\
    \midrule
    ISIC\text{-}2020
        & \checkmark &            &            &            &            & 0.93 & 0.91 & 0.89 \\
        & \checkmark & \checkmark &            &            &            & 0.94 & 0.92 & 0.91 \\
        & \checkmark & \checkmark & \checkmark &            &            & 0.95 & 0.94 & 0.93 \\
        & \checkmark & \checkmark & \checkmark & \checkmark &            & 0.96 & 0.94 & 0.93 \\
        & \checkmark & \checkmark & \checkmark & \checkmark & \checkmark & \textbf{0.97} &
          \textbf{0.95} & \textbf{0.94} \\
    \bottomrule
  \end{tabular}
\end{table}

Table~\ref{table:ablation} quantifies the impact of Deformable Convolution (DC), Scale-Specific Attention (SSA), Multi-Scale Attention (MSA), and Self-Distillation (SD). On PBC, each component improves over the baseline, with additive gains yielding the best accuracy--F1 trade-off. On the highly imbalanced ISIC-2020, all variants use identical class-balanced training, as the unbalanced baseline collapses toward benign prediction. Reduced ablations on C\text{-}NMC and ISIC-2020 show a consistent monotonic trend, where DC and SSA strengthen discrimination, MSA improves cross-scale fusion, and SD provides further refinement, producing gains in AUC, accuracy, and F1. The full configuration performs best across datasets, including under reduced-label settings (Fig.~\ref{fig:subset-pbc}, ~\ref{fig:subset-cnmc}, and ~\ref{fig:subset-isic}), confirming its complementary and generalisable contributions.

\section{Conclusion}

\begin{figure*}[!t]
\centering

\begin{subfigure}[t]{0.32\textwidth}
    \centering
    \includegraphics[width=\linewidth]{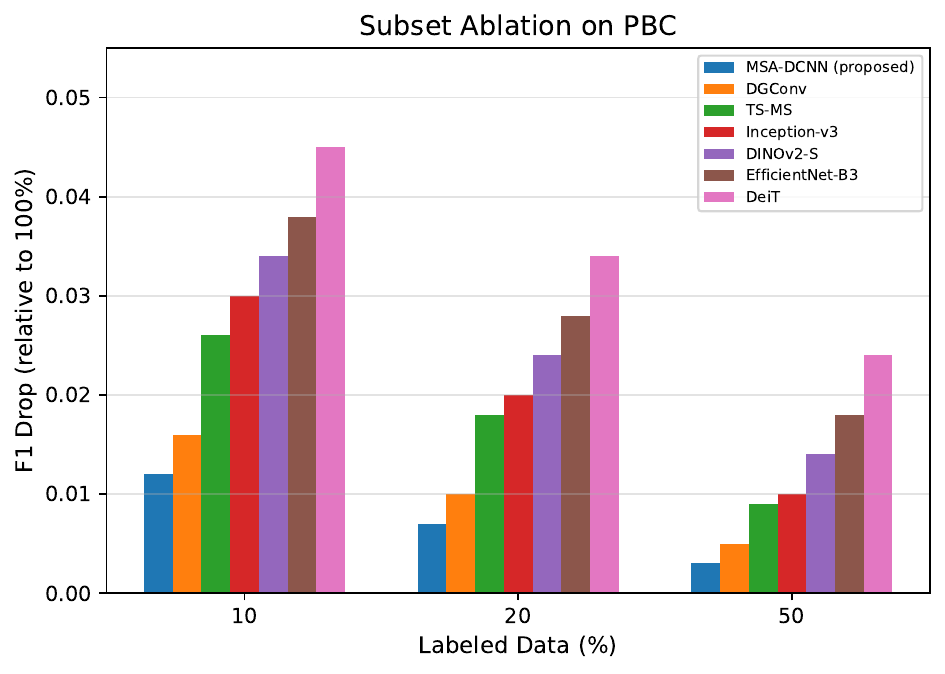}
    \caption{PBC}
    \label{fig:subset-pbc}
\end{subfigure}
\hfill
\begin{subfigure}[t]{0.32\textwidth}
    \centering
    \includegraphics[width=\linewidth]{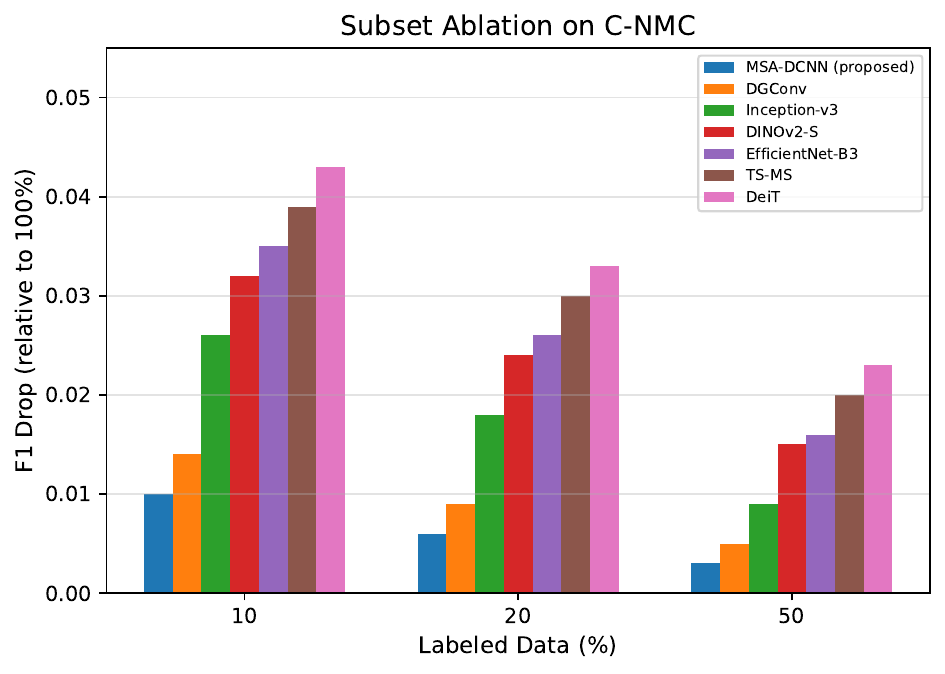}
    \caption{C-NMC}
    \label{fig:subset-cnmc}
\end{subfigure}
\hfill
\begin{subfigure}[t]{0.32\textwidth}
    \centering
    \includegraphics[width=\linewidth]{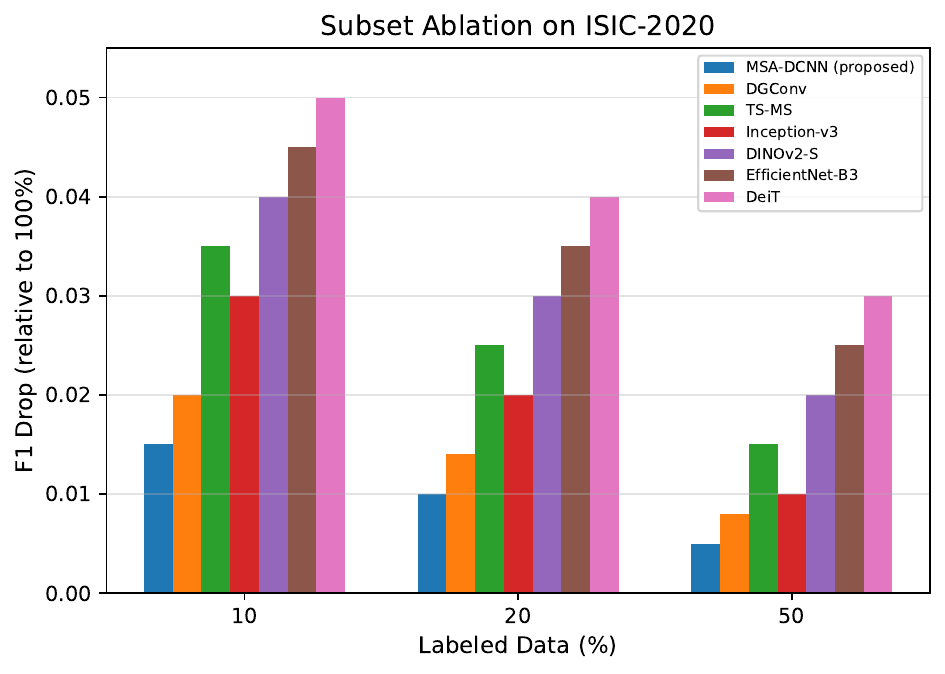}
    \caption{ISIC-2020}
    \label{fig:subset-isic}
\end{subfigure}

\caption{Subset-ablation analysis under reduced labelled-data settings. MSA-DCNN generally exhibits lower F1 degradation across PBC, C-NMC, and ISIC-2020, suggesting improved robustness and label efficiency under limited-supervision settings compared with the evaluated baselines.}
\label{fig:subset-ablation-all}

\end{figure*}

We presented MSA-DCNN, a scale-consistent framework for data-efficient medical image classification that integrates adaptive sampling, scale-specific feature refinement, learned multi-scale attention fusion, and auxiliary self-distillation. Across C-NMC, PBC, ISIC-2020, and an external leukaemia hold-out set, it achieves competitive and often better performance; ablations confirm complementary contributions of DC, SSA, MSA, and SD, with consistent gains under reduced labels (Fig.~\ref{fig:subset-ablation-all}), supporting robustness under scanner, staining, and cohort shift. Future work will explore uncertainty-aware inference, enhanced interpretability, and trustworthy clinical decision support.

\bibliographystyle{splncs04}
\bibliography{mybibliography}

@article{chen2022recent,
  title={Recent advances and clinical applications of deep learning in medical image analysis},
  author={Chen, Xuxin and Wang, Ximin and Zhang, Ke and Fung, Kar-Ming and Thai, Theresa C and Moore, Kathleen and Mannel, Robert S and Liu, Hong and Zheng, Bin and Qiu, Yuchen},
  journal={Medical image analysis},
  volume={79},
  pages={102444},
  year={2022},
  publisher={Elsevier}
}

@article{subba2025computationally,
  title={Computationally optimized brain tumor classification using attention based GoogLeNet-style CNN},
  author={Subba, Anjana Bharati and Sunaniya, Arun Kumar},
  journal={Expert Systems with Applications},
  volume={260},
  pages={125443},
  year={2025},
  publisher={Elsevier}
}

@article{jin2023simplified,
  title={Simplified inception module based Hadamard attention mechanism for medical image classification},
  author={Jin, Yanlin and You, Zhiming and Cai, Ningyin},
  journal={Journal of Computer and Communications},
  volume={11},
  number={6},
  pages={1--18},
  year={2023},
  publisher={Scientific Research Publishing}
}

@article{xu2022isanet,
  title={ISANET: Non-small cell lung cancer classification and detection based on CNN and attention mechanism},
  author={Xu, Zhiwen and Ren, Haijun and Zhou, Wei and Liu, Zhichao},
  journal={Biomedical Signal Processing and Control},
  volume={77},
  pages={103773},
  year={2022},
  publisher={Elsevier}
}

@article{han2024dm,
  title={DM-CNN: Dynamic Multi-scale Convolutional Neural Network with uncertainty quantification for medical image classification},
  author={Han, Qi and Qian, Xin and Xu, Hongxiang and Wu, Kepeng and Meng, Lun and Qiu, Zicheng and Weng, Tengfei and Zhou, Baoping and Gao, Xianqiang},
  journal={Computers in biology and medicine},
  volume={168},
  pages={107758},
  year={2024},
  publisher={Elsevier}
}

@article{khalifa2024ai,
  title={AI in diagnostic imaging: Revolutionising accuracy and efficiency},
  author={Khalifa, Mohamed and Albadawy, Mona},
  journal={Computer Methods and Programs in Biomedicine Update},
  pages={100146},
  year={2024},
  publisher={Elsevier}
}

@misc{obuchowicz2024clinical,
  title={Clinical applications of artificial intelligence in medical imaging and image processing—A review},
  author={Obuchowicz, Rafa{\l} and Strzelecki, Micha{\l} and Pi{\'o}rkowski, Adam},
  journal={Cancers},
  volume={16},
  number={10},
  pages={1870},
  year={2024},
  publisher={MDPI}
}

@article{guo2025artificial,
  title={Artificial intelligence-based automated breast ultrasound radiomics for breast tumor diagnosis and treatment: a narrative review},
  author={Guo, Yinglin and Li, Ning and Song, Chonghui and Yang, Juan and Quan, Yinglan and Zhang, Hongjiang},
  journal={Frontiers in Oncology},
  volume={15},
  pages={1578991},
  year={2025}
}

@article{rayed2024deep,
  title={Deep learning for medical image segmentation: State-of-the-art advancements and challenges},
  author={Rayed, Md Eshmam and Islam, SM Sajibul and Niha, Sadia Islam and Jim, Jamin Rahman and Kabir, Md Mohsin and Mridha, MF},
  journal={Informatics in Medicine Unlocked},
  pages={101504},
  year={2024},
  publisher={Elsevier}
}

@article{mall2023comprehensive,
  title={A comprehensive review of deep neural networks for medical image processing: Recent developments and future opportunities},
  author={Mall, Pawan Kumar and Singh, Pradeep Kumar and Srivastav, Swapnita and Narayan, Vipul and Paprzycki, Marcin and Jaworska, Tatiana and Ganzha, Maria},
  journal={Healthcare Analytics},
  volume={4},
  pages={100216},
  year={2023},
  publisher={Elsevier}
}

@article{mienye2025deep,
  title={Deep convolutional neural networks in medical image analysis: A review},
  author={Mienye, Ibomoiye Domor and Swart, Theo G and Obaido, George and Jordan, Matt and Ilono, Philip},
  journal={Information},
  volume={16},
  number={3},
  pages={195},
  year={2025},
  publisher={Mdpi}
}

@article{yu2021convolutional,
  title={Convolutional neural networks for medical image analysis: state-of-the-art, comparisons, improvement and perspectives},
  author={Yu, Hang and Yang, Laurence T and Zhang, Qingchen and Armstrong, David and Deen, M Jamal},
  journal={Neurocomputing},
  volume={444},
  pages={92--110},
  year={2021},
  publisher={Elsevier}
}

@article{tong2024hybrid,
  title={Hybrid attention mechanism of feature fusion for medical image segmentation},
  author={Tong, Shanshan and Zuo, Zhentao and Liu, Zuxiang and Sun, Dengdi and Zhou, Tiangang},
  journal={IET Image Processing},
  volume={18},
  number={1},
  pages={77--87},
  year={2024},
  publisher={Wiley Online Library}
}

@article{wei2021fine,
  title={Fine-grained image analysis with deep learning: A survey},
  author={Wei, Xiu-Shen and Song, Yi-Zhe and Mac Aodha, Oisin and Wu, Jianxin and Peng, Yuxin and Tang, Jinhui and Yang, Jian and Belongie, Serge},
  journal={IEEE transactions on pattern analysis and machine intelligence},
  volume={44},
  number={12},
  pages={8927--8948},
  year={2021},
  publisher={IEEE}
}

@article{zhang2022rcmnet,
  title={RCMNet: A deep learning model assists CAR-T therapy for leukemia},
  author={Zhang, Ruitao and Han, Xueying and Lei, Zhengyang and Jiang, Chenyao and Gul, Ijaz and Hu, Qiuyue and Zhai, Shiyao and Liu, Hong and Lian, Lijin and Liu, Ying and others},
  journal={Computers in biology and medicine},
  volume={150},
  pages={106084},
  year={2022},
  publisher={Elsevier}
}

@article{kishore2025designing,
  title={Designing Autotuned Student Models by Knowledge Distillation from Self-Supervised Teacher Models},
  author={Kishore, Jaydeep and Mukherjee, Snehasis},
  journal={SN Computer Science},
  volume={6},
  number={5},
  pages={1--16},
  year={2025},
  publisher={Springer}
}

@inproceedings{hussaini2025modified,
  title={Modified CBAM: Sub-block Pooling for Improved Channel and Spatial Attention},
  author={Hussaini, Hamza and Bano, Shahana and Elyan, Eyad and Moreno-Garcia, Carlos Francisco},
  booktitle={Annual Conference on Medical Image Understanding and Analysis},
  pages={116--130},
  year={2025},
  organization={Springer}
}

@article{acevedo2020dataset,
  title={A dataset of microscopic peripheral blood cell images for development of automatic recognition systems},
  author={Acevedo, Andrea and Merino, Anna and Alf{\'e}rez, Santiago and Molina, {\'A}ngel and Bold{\'u}, Laura and Rodellar, Jos{\'e}},
  journal={Data in brief},
  volume={30},
  pages={105474},
  year={2020}
}

@inproceedings{yu2025small,
  title={Small Lesions-aware Bidirectional Multimodal Multiscale Fusion Network for Lung Disease Classification},
  author={Yu, Jianxun and Ge, Ruiquan and Wang, Zhipeng and Yang, Cheng and Lin, Chenyu and Fu, Xianjun and Liu, Jikui and Elazab, Ahmed and Wang, Changmiao},
  booktitle={International Conference on Medical Image Computing and Computer-Assisted Intervention},
  pages={589--598},
  year={2025},
  organization={Springer}
}

@inproceedings{touvron2021training,
  title={Training data-efficient image transformers \& distillation through attention},
  author={Touvron, Hugo and Cord, Matthieu and Douze, Matthijs and Massa, Francisco and Sablayrolles, Alexandre and J{\'e}gou, Herv{\'e}},
  booktitle={International conference on machine learning},
  pages={10347--10357},
  year={2021},
  organization={PMLR}
}

@inproceedings{szegedy2016rethinking,
  title={Rethinking the inception architecture for computer vision},
  author={Szegedy, Christian and Vanhoucke, Vincent and Ioffe, Sergey and Shlens, Jon and Wojna, Zbigniew},
  booktitle={Proceedings of the IEEE conference on computer vision and pattern recognition},
  pages={2818--2826},
  year={2016}
}

@inproceedings{tan2019efficientnet,
  title={Efficientnet: Rethinking model scaling for convolutional neural networks},
  author={Tan, Mingxing and Le, Quoc},
  booktitle={International conference on machine learning},
  pages={6105--6114},
  year={2019},
  organization={PMLR}
}

@article{adepu2023melanoma,
  title={Melanoma classification from dermatoscopy images using knowledge distillation for highly imbalanced data},
  author={Adepu, Anil Kumar and Sahayam, Subin and Jayaraman, Umarani and Arramraju, Rashmika},
  journal={Computers in Biology and Medicine},
  volume={154},
  pages={106571},
  year={2023},
  publisher={Elsevier}
}

@article{gupta2022c,
  title={C-NMC: B-lineage acute lymphoblastic leukaemia: A blood cancer dataset},
  author={Gupta, Ritu and Gehlot, Shiv and Gupta, Anubha},
  journal={Medical Engineering \& Physics},
  volume={103},
  pages={103793},
  year={2022},
  publisher={Elsevier}
}

@inproceedings{li2024semi,
  title={Semi-supervised lymph node metastasis classification with pathology-guided label sharpening and two-streamed multi-scale fusion},
  author={Li, Haoshen and Wang, Yirui and Zhu, Jie and Guo, Dazhou and Yu, Qinji and Yan, Ke and Lu, Le and Ye, Xianghua and Zhang, Li and Wang, Qifeng and others},
  booktitle={International Conference on Medical Image Computing and Computer-Assisted Intervention},
  pages={623--633},
  year={2024},
  organization={Springer}
}

@inproceedings{ju2024universal,
  title={Universal Semi-supervised Learning for Medical Image Classification},
  author={Ju, Lie and Wu, Yicheng and Feng, Wei and Yu, Zhen and Wang, Lin and Zhu, Zhuoting and Ge, Zongyuan},
  booktitle={International Conference on Medical Image Computing and Computer-Assisted Intervention},
  pages={355--365},
  year={2024},
  organization={Springer}
}

@inproceedings{kunanbayev2024training,
  title={Training vit with limited data for Alzheimer’s disease classification: An empirical study},
  author={Kunanbayev, Kassymzhomart and Shen, Vyacheslav and Kim, Dae-Shik},
  booktitle={International Conference on Medical Image Computing and Computer-Assisted Intervention},
  pages={334--343},
  year={2024},
  organization={Springer}
}

@inproceedings{dai2017deformable,
  title={Deformable convolutional networks},
  author={Dai, Jifeng and Qi, Haozhi and Xiong, Yuwen and Li, Yi and Zhang, Guodong and Hu, Han and Wei, Yichen},
  booktitle={Proceedings of the IEEE international conference on computer vision},
  pages={764--773},
  year={2017}
}

@article{oquab2023dinov2,
  title={Dinov2: Learning robust visual features without supervision},
  author={Oquab, Maxime and Darcet, Timoth{\'e}e and Moutakanni, Th{\'e}o and Vo, Huy and Szafraniec, Marc and Khalidov, Vasil and Fernandez, Pierre and Haziza, Daniel and Massa, Francisco and El-Nouby, Alaaeldin and others},
  journal={arXiv preprint arXiv:2304.07193},
  year={2023}
}

@inproceedings{gong2021deformable,
  title={Deformable gabor feature networks for biomedical image classification},
  author={Gong, Xuan and Xia, Xin and Zhu, Wentao and Zhang, Baochang and Doermann, David and Zhuo, Li'an},
  booktitle={Proceedings of the IEEE/CVF Winter Conference on applications of computer vision},
  pages={4004--4012},
  year={2021}
}
\end{document}